\documentclass{article}
\usepackage{spconf,amsmath,graphicx, booktabs}
\usepackage{algorithm, algorithmic, multirow, amssymb, array, makecell}
\usepackage[dvipsnames]{xcolor}
\usepackage{url}
\usepackage{adjustbox}

\newcommand{\mycomment}[1]{}

\title{Multimodal  Attention Merging For Improved Speech Recognition and Audio Event Classification}
\name{\begin{tabular}{@{}c@{}}Anirudh S. Sundar\textsuperscript{\rm 1,2,$\dagger$}\sthanks{Corresponding authors \textit{asundar34@gatech.edu} and  \textit{phanisn@amazon.com}, $\dagger$ Work performed as an intern at Amazon}\qquad  Chao-Han Huck Yang\textsuperscript{\rm 2} \qquad David M. Chan\textsuperscript{\rm 2, 3}  \\ \textit{Shalini Ghosh}\textsuperscript{\rm 2} \qquad \textit{Venkatesh Ravichandran}\textsuperscript{\rm 2} \qquad \textit{Phani Sankar Nidadavolu}\textsuperscript{\rm 2*}\end{tabular}}
\address{\textsuperscript{\rm 1}Georgia Institute of Technology  \; \textsuperscript{\rm 2}Amazon Alexa AI \; \textsuperscript{\rm 3}University of California, Berkeley}
\begin{document}
\ninept
\maketitle
\begin{abstract}
Training large foundation models using self-supervised objectives on unlabeled data, followed by fine-tuning on downstream tasks, has emerged as a standard procedure. Unfortunately, the efficacy of this approach is often constrained by both limited fine-tuning compute and scarcity in labeled downstream data.  We introduce Multimodal Attention Merging (MAM), an attempt that facilitates direct knowledge transfer from attention matrices of models rooted in high-resource modalities, text and images, to those in resource-constrained domains, speech and audio, employing a \textit{zero-shot paradigm}. MAM reduces the relative Word Error Rate (WER) of an Automatic Speech Recognition (ASR) model by up to 6.70\%, and relative classification error of an Audio Event Classification (AEC) model by 10.63\%. In cases where some data/compute is available, we present Learnable-MAM, a data-driven approach to merging attention matrices, resulting in a further 2.90\% relative reduction in WER for ASR and 18.42\% relative reduction in AEC compared to fine-tuning without model merging. 
\end{abstract}
\begin{keywords}
Knowledge transfer, cross-modal adaptation, speech recognition, and acoustic modeling
\end{keywords}
\section{Introduction}
\label{sec:intro}

Current approaches in deep learning train large foundation models using task-agnostic self-supervised objectives on unlabeled data \cite{brown2020language, hsu2021hubert, dosovitskiy2020image}. The models are then fine-tuned on downstream tasks, utilizing task-specific inputs and labels.
As foundation models increase in size, fine-tuning is hamstrung by limitations in compute. Moreover, scarcity of task-specific data compounds this challenge, particularly for relatively lower-resource modalities like speech or audio when compared to more widely studied modalities such as text or images.

Prior research \cite{lu2021pretrained, yang2021voice2series, gu2022can, reid2022can} has demonstrated transferability of the Transformer \cite{vaswani2017attention} across modalities with minimal or no fine-tuning of the self-attention mechanism. \cite{lu2021pretrained} shows that text pre-training is sufficient to learn modality agnostic properties of sequences by demonstrating the success of frozen text pre-trained Transformers in image classification and protein fold prediction without self-attention fine-tuning. \cite{ri2022pretraining} shows that cross-modality capabilities in Transformers arises from the transfer of knowledge in the form of position-aware context. \cite{schwettmann2023multimodal} and \cite{goh2021multimodal} provide evidence for multimodal neurons in the Transformer and demonstrate that modality transfer happens in intermediate layers.

Motivated by these works, we present Multimodal Attention Merging (MAM) to investigate the possibility of transferring knowledge from models trained on high-resource modalities such as text and images to models trained on relatively low-resource modalities such as speech and audio. Given abundant textual and visual data, self-supervised pre-training using objectives such as Masked Language Modeling and Masked Patch Prediction utilize the Transformer to learn generalized representations of natural language text and images. Chiang and Lee \cite{chiang2022transferability} indicate that using Masked Language Modeling allows self-attention to capture explicit and implicit dependencies at the token level, which are modality agnostic. Therefore, through MAM we investigate whether these textual and visual parameter-space representations generalize to speech and audio. Through a systematic interpolation of attention matrices from models trained on high-resource modalities (e.g., BERT \cite{devlin2019bert}, Vision Transformer \cite{dosovitskiy2020image}) MAM demonstrates an improvement in performance of models trained on low-resource modalities (e.g., HuBERT \cite{hsu2021hubert}, BEATs \cite{chen2022beats}) on Automatic Speech Recognition~(ASR) and Audio Event Classification (AEC). Our contributions are:




\begin{figure}[t]
    \centering
    \includegraphics[width=\columnwidth]{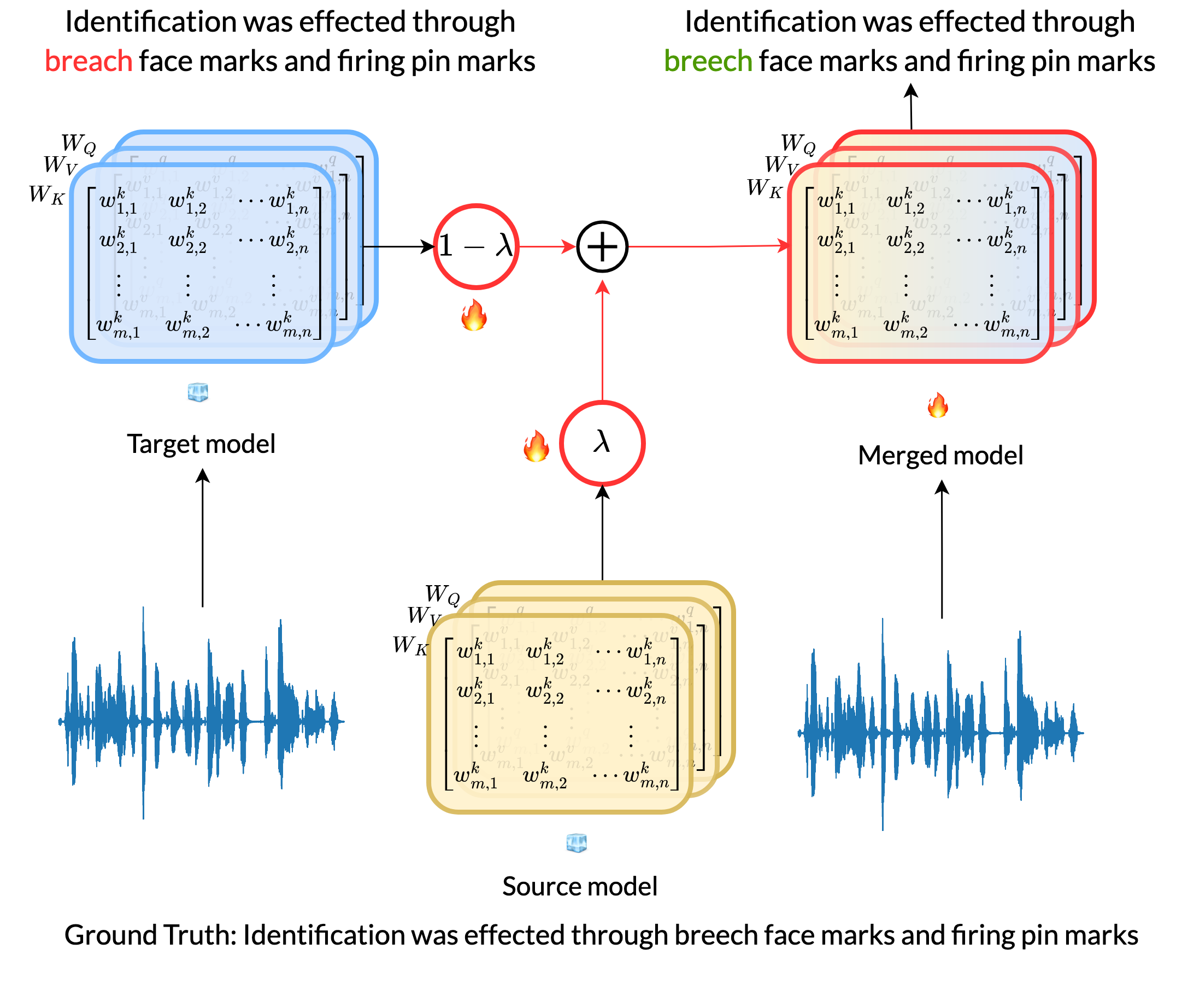}
    \caption{Learnable Multimodal Attention Merging. The merged model correctly transcribes \textit{breech}. \includegraphics[height=9pt]{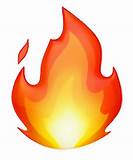} modules are fine-tuned (trainable) while \includegraphics[height=9pt]{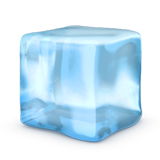} are frozen (non-trainable). In our initial approach, Multimodal Attention Merging, the interpolation factor $\lambda$ is frozen (non-trainable).}
    \vspace{-15pt}
    \label{fig:model_merging}
\end{figure}

\begin{itemize}
    \item In contrast to prior research \cite{sung_empirical_2023, wortsman2022fi, choshen2022fusing, singh2020model, wortsman2022model}, we show attention-merging on models from different tasks, modalities, and initializations. To the best of our knowledge, we are the first to extend attention-merging to sequence-to-sequence tasks, specifically ASR. 

    \item Introducing Multimodal Attention Merging~(MAM), we lower HuBERT's relative Word Error Rate (WER) on LJ Speech by 6.70\% and VCTK by 1.80\%, also decreasing BEATs' relative classification error on ESC-50 by 10.63\%, without additional fine-tuning (Section \ref{subsec:att_int}).

    \item We present a data-driven approach employing the Sliced Wasserstein Distance \cite{kolouri2019generalized} between intermediate hidden-layer representations to merge specific attention layers, reducing the need for empirical experimentation (Section \ref{subsec:expts_zs}).
    
    \item In the case where some data/compute is available, we present Learnable-MAM (L-MAM), by learning the interpolation factor during fine-tuning. L-MAM yields a 2.70\% and 2.90\% relative reduction in WER on LJ Speech and VCTK and a 18.42\% relative reduction in classification error on ESC-50 compared to regular fine-tuning (Section \ref{subsec:learnable_int}). 
    
\end{itemize}

\section{Related Work}
\label{sec:related_work}


Prior work in multimodal merging includes OTKGE \cite{cao2022otkge}, that uses Optimal Transport to align structural knowledge, linguistic information, and image embeddings in knowledge graphs. However, they do not extend their approach to merging model weights and focus on embeddings instead. Voice2Series \cite{yang2021voice2series} and Frozen Pretrained Transformer \cite{lu2021pretrained} demonstrate knowledge transfer across modalities through frozen self-attention weights. While both works study the transferability of self-attention across modalities, they stop short of merging models trained from different modalities and do not address sequence-to-sequence tasks such as ASR. In contrast, works such as Fisher merging \cite{matena2022merging}, \textit{local fine-tuning} \cite{wortsman2022fi}, \textsc{Model Soups} \cite{wortsman2022model}, DMC \cite{zhang:wacv:2020}, \textsc{AdapterSoup} \cite{chronopoulou2023adaptersoup}, MLM \cite{mahadevan:ecml:2018} discuss model merging but do not consider the multimodal scenario. Perhaps the closest work to ours is Multimodal Model Merging \cite{sung_empirical_2023}, an empirical study of merging vision and text models for combined vision-language tasks such as Visual Question Answering \cite{antol2015vqa} and image-text retrieval \cite{lin2014microsoft}. The approach studies the use of simple interpolation, \textsc{RegMean} \cite{sung_empirical_2023}, and Task Vectors \cite{ilharco2022editing} to determine the best model merging approach. However, they require contrastive model alignment, use a shared seed pre-training phase to initialize models prior to merging, and address joint vision+language tasks. In contrast, MAM merges off-the-shelf models from different modalities without constraints on pre-training tasks or weight initialization. Through L-MAM, we also present an approach to learn the interpolation factor in the case where limited data/compute is available, reducing the requirement on empirical experimentation.

\section{Method}
\label{sec:method}

MAM seeks to determine if the Transformer \cite{vaswani2017attention} attention mechanism generalizes across modalities. It does so by exploring the transferability of parameter-space sequence representations of Transformers pre-trained on high-resource modalities (text or vision) to those trained on relatively low-resource modalities (speech or audio). For convenience, we refer to the high-resource and low-resource pre-trained models as the \textit{Source Model} and \textit{Target Model} respectively. Figure~\ref{fig:model_merging} presents an outline of attention merging. It is important to note that all our approaches require the source and target models to have the same number of attention layers and hidden layer sizes. 

We apply attention merging to two tasks: Automatic Speech Recognition (ASR) and Audio Event Classification (AEC). ASR transcribes human speech to text, while AEC identifies real-life events (e.g. barking, thunderstorm, whistling) from audio clips. 

We use three main approaches to demonstrate MAM: Attention Interpolation, Layer-wise Attention Interpolation, and Attention-Merging with Learnable Interpolation. 

\subsection{Attention Interpolation}
\label{subsec:att_int}
MAM with attention interpolation uses a convex combination of the source and target models. Using an interpolation factor $\lambda \in [0,1]$, we merge Query, Key, and Value matrices ($W_Q, W_K, W_V$) across all layers $\mathcal{L}$ in the attention computation. For source and target models $M_s$ and $M_t$, the merged model $M_{\textsf{merge}}$'s Query, Key, and Value matrices for each layer $i$ are defined in Equation \ref{eq:model_interpolation}. $M_{\textsf{merge}}$ is applied to the same downstream task as $M_t$.

\begin{align}
    W_{Q_i,K_i,V_i}^{M_{\textsf{merge}}} = \lambda \cdot W_{Q_i, K_i, V_i}^{M_s} + &(1-\lambda) \cdot W_{Q_i,K_i,V_i }^{M_t} \label{eq:model_interpolation}
\end{align}


\subsection{Layer-wise Attention Interpolation}
\label{subsec:att_lyr_int}
Evidenced by research categorizing layers by importance~\cite{zhang2022all}, we experiment with merging a subset of layers $\ell$ from the set of all layers $\mathcal{L}$ ($|\ell| < |\mathcal{L}|$) and present the modified approach in Equation \ref{eq:model_layer_interpolation}. We are interested in identifying whether merging all layers is necessary, or if merging a subset yields better generalization. 
\begin{alignat}{2}
    W_{Q_i,K_i,V_i}^{M_{\textsf{merge}}} = \lambda \cdot W_{Q_i, K_i, V_i}^{M_s} + (1-\lambda) \cdot W_{Q_i,K_i,V_i }^{M_t} && \label{eq:model_layer_interpolation} \\ 
    \lambda = 0 \; \forall{i} \notin \ell && \notag
\end{alignat}


\subsection{Attention-Merging with Learnable Interpolation}
\label{subsec:learnable_int}
Finally, we learn the interpolation factor $\lambda$ for individual downstream tasks. In this setting, $\lambda$ is optimized concurrently with model weights $M_t$ during fine-tuning. We omit optimizing $M_s$ as it corresponds to a different modality from the downstream task. In contrast to the previous two approaches, we do not impose the requirement for utilizing uniform $\lambda$ across layers. Instead, we view learning $\lambda$ as a gate that enables flexible information transmission from the high-resource to the low-resource modality and describe this technique in Equation \ref{eq:model_interpolation_learnable}.

\begin{alignat}{2}
    W_{Q_i,K_i,V_i}^{M_{\textsf{merge}}} = \lambda_i \cdot W_{Q_i, K_i, V_i}^{M_s} + (1-\lambda_i) \cdot W_{Q_i,K_i,V_i }^{M_t} && \label{eq:model_interpolation_learnable}
\end{alignat}

\section{Experiments}
\label{sec:Experiments}
For ASR, we merge publicly available HuBERT-large \cite{hsu2021hubert} and BERT-large-uncased \cite{devlin2019bert} models. Both models have 24 attention layers and a width of 1024, totaling 300 million parameters. HuBERT-large is an encoder model pre-trained on 60,000 hours of Libri-Light \cite{librilight} for speech representation learning and fine-tuned  with CTC on 960h of Librispeech \cite{panayotov2015librispeech} for ASR. BERT-large-uncased is pre-trained via Masked Language Modeling on the Wikipedia corpus and Bookcorpus. We evaluate merged models on LJ Speech \cite{ljspeech17}, a single-speaker dataset of 13,100 audio clips from non-fiction books totaling 24 hours, and VCTK \cite{Yamagishi2019CSTRVC}, a multi-speaker dataset with 43,000 audio clips totaling 44 hours. VCTK features diverse speakers from regions like England, Scotland, and America, reading texts chosen for comprehensive contextual and phonetic coverage.

For AEC, we merge BEATs \cite{chen2022beats} and Vision Transformer (ViT)~\cite{dosovitskiy2020image} containing 12 attention layers and hidden size of 768, totaling 90 million parameters. BEATs is pretrained on AudioSet~\cite{45857} for audio representation learning while ViT is pretrained on ImageNet~\cite{5206848}. Evaluation focuses on ESC-50 \cite{piczak2015dataset}, which comprises 2000 5-second environmental audio recordings across 50 classes. 

\mycomment{
\begin{table}[t]
\begin{adjustbox}{width=0.48\textwidth}
    \centering
    \begin{tabular}{c|c|c|c|c}
    \toprule 
        Task & \makecell{Source \\ Model} & \makecell{Target \\ Model} & Layers & \#Params \\
        \midrule 
        \makecell{Automatic Speech Recognition } & BERT & HuBERT & 24 & 300M \\
        \makecell{Audio Classification} & ViT & BEATs & 12 & 90M \\
        \bottomrule
    \end{tabular}
        \end{adjustbox}
    \caption{Summary of source and target model architectures}
    \label{tab:model_summary}
\end{table}
}

\subsection{Case Study 1: Zero-Shot Experiments}
\label{subsec:expts_zs}
The first study evaluates zero-shot performance of merged models on downstream datasets. We use a dev set to select the interpolation $\lambda$ and report results on a held-out test~set. 

\textbf{Attention-Interpolation.}
Table \ref{tab:MM_Entire_LJS_VCTK} contains the results for attention merging of entire models as described in Section \ref{subsec:att_int} for different interpolation factors $\lambda$. We merge source and target models following the strategy outlined in Equation \ref{eq:model_interpolation}. As an additional baseline, we also merge target models with source models sampled from random noise. We experiment with three types of noise - sampling noise from source model parameters $\mathcal{N}$(Source), target model parameters $\mathcal{N}$(Target), and a standard normal distribution $\mathcal{N}$(0,1). For $\mathcal{N}$(Source) and $\mathcal{N}$(Target), we draw samples with the same mean and variance as the parameters of the source and target models respectively. The samples are then used to construct attention matrices with the same dimensions as the Target model. The results in Table~\ref{tab:MM_Entire_LJS_VCTK} underscore the clear advantage of merging attention matrices from the source model. Conversely, merging with noise substantially diminishes performance, especially at higher interpolation factors. 

\begin{table}[t]
    \centering
    \begin{tabular}{c|c|c|c|c}
        \toprule 
        $\lambda$ & \makecell{\small{Source}} & \makecell{$\mathcal{N}$(\small{Source})} & $\mathcal{N}$(\small{Target}) & $\mathcal{N}(0,1)$  \\ \midrule 
        \midrule 
        \multicolumn{5}{l}{\normalsize{LJ Speech - WER(\%) Source: BERT   Target: HuBERT}} \\ \midrule 
        0.00 & 9.25 & -     &    - & -    \\ 
        0.05 & 9.11 & 9.78  & 9.26 & 10.57 \\ 
        0.10 & \textbf{9.06} & 32.43 & 10.06 & 74.03 \\ 
        0.15 & 9.20 & 98.30 & 18.13 & 99.95 \\
        0.20 & 9.62 & 99.99 & 75.93 & 100.00 \\
        0.25 & 11.12 & 100.00 & 100.00 & 100.00 \\ \midrule 
        \midrule 
        \multicolumn{5}{l}{\normalsize{VCTK - WER(\%) Source: BERT   Target: HuBERT}} \\ \midrule 
        0.00 & 5.58 & -    &  -   &  -   \\      
        0.05 & \textbf{5.48} & 5.84  & 5.54 & 6.26 \\ 
        0.10 & 5.55 & 22.57 & 6.17 & 54.53 \\ 
        0.15 & 5.85 & 91.93 & 12.81 & 99.83 \\
        0.20 & 6.45 & 99.42 & 58.57 & 100.00 \\
        0.25 & 8.50 & 99.96 & 97.15 & 100.00 \\ \midrule 
        \midrule 
        \multicolumn{5}{l}{\normalsize{ESC-50 - Error(\%) Source: ViT   Target: BEATs}} \\ \midrule 
        0.00 & 11.75 &   -   & -    &  - \\
        0.10 & \textbf{11.50} & 11.76 & 83.40 & 98.25  \\
        0.15 & 11.75 & 12.44 & 98.00 & 96.75  \\
        0.20 & 17.50 & 14.36 & 98.00 & 97.00 \\
        0.25 & 16.75 & 19.12 & 98.00 & 98.50 \\
        0.30 & 33.00 & 26.82 & 98.00 & 98.50 \\
        \bottomrule 
    \end{tabular}
    \caption{Error(\%) for interpolation of source and target models with factor $\lambda$. $\mathcal{N}$(Source) and $\mathcal{N}$(Target) are noise sampled with the same mean and variance as parameters of source and target models.}
    \label{tab:MM_Entire_LJS_VCTK}
    \vspace{-15pt}
\end{table}


\textbf{Interpolation of subset of layers.}
Table \ref{tab:LJS_VCTK_ZS_subset} summarizes the results for merging a subset of layers. We experimented with merging blocks of different sizes and different interpolation but for the sake of brevity present results on blocks of 8 layers for ASR and 4 or 8 layers for AEC with the best performing $\lambda$ selected using a dev set, which was 0.25 for LJ Speech, 0.05 for VCTK, and 0.1 for ESC-50. For ASR, we identified that merging layers 12-19 resulted in the lowest WER while it was layers 4-11 for AEC. 

To determine whether a better subset exists, we employed a data-driven strategy to pinpoint the most suitable layers to merge. We used audio snippets and corresponding text transcripts from 10\% of the training set of LJ Speech and VCTK. HuBERT encoded the audio, while BERT encoded the transcripts. For each sample, hidden representations at the output of each attention block were extracted from every layer of both networks. These hidden representations were averaged by sequence length, resulting in a 1024-dimensional vector per layer for each sample. Similarity between an audio-text pair's HuBERT and BERT hidden representations for each layer was computed via Euclidean distance, the inner product, and motivated by prior work that compares hidden representations in speech models \cite{chen2023estimate}, the Sliced Wasserstein Distance (SWD) \cite{kolouri2019generalized}. Sorting layers by similarity, the top-k most similar layers were merged. Results for distinct k values and distance metrics are presented in Table~\ref{tab:layer_selection}. For LJ Speech, we find that merging top 6 layers identified using SWD yields best results while it is the top 10 layers for VCTK. Comparing Tables \ref{tab:LJS_VCTK_ZS_subset} and \ref{tab:layer_selection}, we notice that merging continuous blocks of layers performs better or almost identical to the data-driven strategy. In contrast, the data-driven strategy does not require extensive experimentation. We did not perform data-driven experiments for AEC due to the absence of paired images for audio snippets in ESC-50.

\begin{table}[t]
\centering
\begin{tabular}{c|c|c||c|c}
\toprule
\makecell{Layers \\ Merged} &  \makecell{LJ Speech \\ $\lambda=0.25$} & \makecell{VCTK \\ $\lambda=0.05$} & \makecell{Layers \\ Merged} & \makecell{ESC-50 \\ $\lambda=0.1$}  \\ \midrule
0 - 7  & 9.33 & 5.53 & 0 - 3 & 13.00  \\
4 - 11  & 9.59 & 5.55 & 4 - 7 & 12.50 \\
8 - 15  & 9.28 & 5.54 & 8 - 11 & 12.00 \\
12 - 19  & \textbf{8.63} & \textbf{5.52} & 0 - 7 & 11.75  \\
16 - 23  & 10.07 & 5.54 & 4 - 11 & \textbf{10.50} \\ \midrule 
None & \multicolumn{1}{c|}{9.25} & \multicolumn{1}{c||}{5.58} & None & \multicolumn{1}{c}{11.75}  \\
\bottomrule
\end{tabular}
\caption{Model Merging on a subset of layers - WER(\%) on LJ Speech and VCTK and Error(\%) on ESC-50}
\label{tab:LJS_VCTK_ZS_subset}
\vspace{-15pt}
\end{table}

\begin{table}[h]
    \centering
    \begin{tabular}{c|c|c|c|c}
    \toprule 
        $\lambda$ & k & \makecell{Euclidean \\ Distance} & \makecell{SWD} & \makecell{Inner \\Product}  \\ \midrule 
        \midrule 
         \multicolumn{5}{c}{LJ Speech} \\ \midrule 
        \multirow{3}*{0.1} & 4 & 9.18 & 8.74 & 9.28 \\
         &  6 & 8.87 & \textbf{8.73} & 9.38 \\
         & 8 & 9.19 & 8.92 & 9.32 \\ 
        \midrule 
        \multicolumn{5}{c}{VCTK} \\ \midrule 
        \multirow{3}*{0.05} & 6 & 5.53 & 5.55 & 5.54 \\
        & 8 & \textbf{5.51} & 5.52 & 5.52 \\
        & 10 & \textbf{5.51} & \textbf{5.51} & 5.52 \\ 
        \bottomrule
    \end{tabular}
    \caption{Comparison of selecting Top-k most similar layers based on different distance metrics, WER(\%) on the ASR task}
    \label{tab:layer_selection}
\end{table}

\begin{table*}[t!]
    \centering
    \begin{tabular}{c|p{8cm}|p{8cm}}
    \toprule 
         &  Baseline & MAM  \\ \midrule 
         & \multicolumn{2}{l}{LJ Speech} \\ \midrule 
        1 & the commission believes that the motorcade \textcolor{red}{rout} selected by agent lawson upon the advice of agent in charge sorrels & the commission believes that the motorcade \textcolor{ForestGreen}{route} selected by agent lawson upon the advice of agent in charge sorrels \\
        2 & here a couple of \textcolor{red}{pye men} had been selling their wares the basket of one of them which was raised upon a four legged stool was upset & here a couple of \textcolor{ForestGreen}{piemen} had been selling their wares the basket of one of them which was raised upon a four legged stool was upset \\
        3 & besides his employers a jeweler named humphreys was in the swim at whose shop in red lion square was discovered a quantity of \textcolor{red}{bass} gold & besides his employers a jeweler named humphreys was in the swim at whose shop in red lion square was discovered a quantity of \textcolor{ForestGreen}{base} gold \\
        4 & \textcolor{red}{aproximately} thirty to forty five seconds after oswalds lunch room encounter with baker and \textcolor{red}{truley} & \textcolor{ForestGreen}{approximately} thirty to forty five seconds after oswalds lunchroom encounter with baker and \textcolor{ForestGreen}{truly}\\
        \midrule 
        & \multicolumn{2}{l}{VCTK} \\ \midrule 
        5 & on the contrary they \textcolor{red}{stant togain} & on the contrary they \textcolor{ForestGreen}{stand to gain} \\
        6 & flanke gordon simpson may also feature in that \textcolor{red}{much} & flanke gordon simpson may also feature in that \textcolor{ForestGreen}{match} \\
        7 & the marshal at the \textcolor{red}{tern} was great & the marshal at the \textcolor{ForestGreen}{turn} was great\\
        8 & they were behind the \textcolor{red}{field} & they were behind the \textcolor{ForestGreen}{wheel} \\
        \bottomrule
    \end{tabular}
    \caption{Analysis of improvements on LJ Speech and VCTK}
    \label{tab:error_analysis}
    \vspace{-15pt}
\end{table*}

\vspace{-15pt}
\subsection{Case Study 2: Fine-Tuning Experiments}
While previous results were based on zero-shot evaluation, we also examined fine-tuned performance. A comparison of different approaches is provided in Table~\ref{tab:ablation_final}. The top row represents the baseline of downstream evaluation without attention merging or fine-tuning. By employing MAM, we improve the original model's performance without additional fine-tuning. We report the better of Attention Interpolation (\ref{subsec:att_int}) and Layer-wise Attention Interpolation (\ref{subsec:att_lyr_int}) as the best performing MAM approach. Another baseline involves fine-tuning models on target data alone (third row), with results averaged over 3 runs. We fine-tune using AdamW \cite{loshchilov2019decoupled} using a learning rate of 1e-5 for ASR and 1e-3 for AEC. We fine-tune for 2 epochs on LJ Speech, 1 epoch for VCTK, and 3 epochs for ESC-50 with a batch size of 32. For ASR, we fine-tune HuBERT using CTC Loss.  Fine-tuning on target data outperforms MAM for LJ Speech and ESC-50, while MAM is the better approach for VCTK. We speculate that the similarity in distribution between VCTK (a larger, multi-speaker dataset) and the pre-training data (Libri-light and Librispeech) limits the efficacy of fine-tuning while the single-speaker nature of LJ Speech allows fine-tuning to adapt better to the dataset.

In the situation some data/compute is available, we explore fine-tuning based approaches. We notice improved performance when combining MAM and fine-tuning as compared to using either approach independently. We utilize the best performing model obtained through model-merging (Tables \ref{tab:MM_Entire_LJS_VCTK} and \ref{tab:LJS_VCTK_ZS_subset}) followed by fine-tuning, shown in the fourth row of Table \ref{tab:ablation_final}. Our top-performing model emerges when jointly learning the interpolation factor and model weights during fine-tuning, as shown in the last row of Table \ref{tab:ablation_final}. With L-MAM followed by fine-tuning, we achieve a 2.70\% relative WER reduction on LJ Speech, a 2.90\% reduction on VCTK, and an 18.42\% reduction in classification error on ESC-50 compared to the regular fine-tuning baseline.

\begin{table}[b!]
    \centering
    \begin{tabular}{c|c|c|c|c|c}
    \toprule 
        \rotatebox{90}{MAM} & \rotatebox{90}{FT} & \rotatebox{90}{L-MAM} &   \makecell{WER (\%) \\ LJ Speech} & \makecell{WER (\%) \\ VCTK} & \makecell{ Error (\%) \\ ESC-50}  \\ \midrule
         -  &   -       &  -    &     9.25  & 5.58 & 11.75 \\ 
          \checkmark & - &   -   &  8.63  & 5.48 & 10.50 \\  
          - & \checkmark &    -   &  8.52 & 5.52 & 9.50   \\
          \checkmark & \checkmark & - & 8.40 & 5.44 & \textbf{7.75} \\
          - & \checkmark & \checkmark & \textbf{8.29} & \textbf{5.36} & \textbf{7.75} \\ \bottomrule
    \end{tabular}
    \caption{Ablation study to compare MAM, Fine-Tuning (FT), and Learnable-MAM for ASR and AEC}
    \label{tab:ablation_final}
\end{table}

\subsection{Analysis of Improvements}
\label{sec:improvements}
We now analyze the improvements observed on the ASR task. Table~\ref{tab:error_analysis} contrasts transcriptions from the baseline (without attention merging or fine-tuning) with those from MAM. MAM improves transcriptions for homophonic words, exemplified by `route', `piemen', and `base' in LJ Speech. VCTK, encompassing various speakers, contains non-standard pronunciations, wherein MAM successfully transcribes ambiguous words accurately. Instances include `stand', `match', `turn', and `wheel', as shown in rows 5 to 8 of Table \ref{tab:error_analysis}. We further categorize improvements based on character insertion, substitution, or deletion in Table \ref{tab:types_of_errors}. While LJ Speech improvements primarily stem from insertion and substitutions, VCTK shows a more balanced distribution among these types of errors.

\begin{table}[ht!]
    \centering
    \begin{tabular}{c|c|c|c}
    \toprule 
        Improvement & Insertion & Substitution & Deletion  \\ \midrule 
        LJ Speech & 61.89 \% & 35.13 \% & 2.98 \% \\
        VCTK & 36.36 \% & 28.28 \% & 35.35 \% \\
        \bottomrule 
    \end{tabular}
    \caption{Categorization of Improvements}
    \label{tab:types_of_errors}
\end{table}

\vspace{-15pt}
\section{Conclusion}
This paper introduces Multimodal Attention Merging (MAM), which transfers knowledge from attention matrices of high-resource modality models (e.g., text and images) to low-resource ones (e.g., speech and audio). MAM improves zero-shot performance of HuBERT (ASR) and BEATs (Audio Event Classification) by merging attention matrices with BERT and Vision Transformer. Additionally, Learnable-MAM (L-MAM) jointly learns interpolation and model weights during fine-tuning, achieving up to 2.90\% relative WER reduction and 18.42\% classification error reduction on ASR and Audio Classification compared to regular fine-tuning, respectively.

While MAM and L-MAM  demonstrate improvements, our work is nascent and we believe there are several avenues for future work warranting investigation. Future work could address merging larger, billion-parameter models representing the state of the art. Developing methods to merge models of differing architectures and merging three or more modalities are other dimensions to extend this work. While we merge attention matrices, future work could address the problem from a parameter efficiency perspective by merging adapter modules \cite{houlsby2019parameter} or low-rank representations \cite{hu2021lora}. Finally, we believe that equipping merged models with cross-modality capabilities to build a generalized multi-task architecture holds promise and urge future work in this direction. 

\clearpage
\small
\bibliographystyle{IEEEbib}
\bibliography{strings,refs}

\end{document}